\documentclass[10pt,twocolumn,letterpaper]{article}

\usepackage[cvprfinal]{cvpr}   

\usepackage{booktabs}
\usepackage{array}
\usepackage{url}
\usepackage{float}
\usepackage{natbib}
\usepackage[utf8]{inputenc} % allow utf-8 input
\usepackage[T1]{fontenc}    % use 8-bit T1 fonts         % simple URL typesetting
\usepackage{booktabs}       % professional-quality tables
\usepackage{amsfonts}       % blackboard math symbols
\usepackage{nicefrac}       % compact symbols for 1/2, etc.
\usepackage{microtype}      % microtypography
\usepackage{xcolor}         % colors
\usepackage{amsmath}
\usepackage{graphicx}
\usepackage{subcaption}
\usepackage{amsmath}  % 数学环境
\usepackage{amssymb}  % 数学符号
\usepackage{bm}       % 加粗数学符号
\usepackage{enumitem}
\usepackage{multirow}
\usepackage{wasysym} % For checkmarks
\usepackage{makecell}% To produce the REVIEW version
\definecolor{cvprblue}{rgb}{0.21,0.49,0.74}
\usepackage[pagebackref,breaklinks,colorlinks,allcolors=cvprblue]{hyperref}

\def\paperID{12487} % *** Enter the Paper ID here
\def\confName{CVPR}
\def\confYear{2026}

\title{Head-wise Adaptive Rotary Positional Encoding for Fine-Grained Image Generation}

\author{Jiaye Li$^{1,2,3,\star}$ \quad Baoyou Chen$^{1,2,3,\star}$ \quad Hui Li$^{1,2,3}$ \quad 
        Zilong Dong$^{4}$ \quad Jingdong Wang$^{5}$ \quad Siyu Zhu$^{1,2,3,\dagger}$\\
        $^{1}$Fudan University, $^{2}$Shanghai Innovation Institute, $^{3}$Shanghai Academy of AI for Science\\
        $^{4}$Alibaba Group, $^{5}$Baidu Inc.\\
}
\begin{document}
\maketitle

\begingroup
\renewcommand{\thefootnote}{\(\star\)}
\footnotetext{Equally contributed to this work}
\endgroup

\begin{figure*}[htbp]
\centering
% \fbox{\rule{0pt}{3cm}\rule{5cm}{0pt}} % 5cm宽，3cm高的空图
\includegraphics[width=1.0\textwidth]{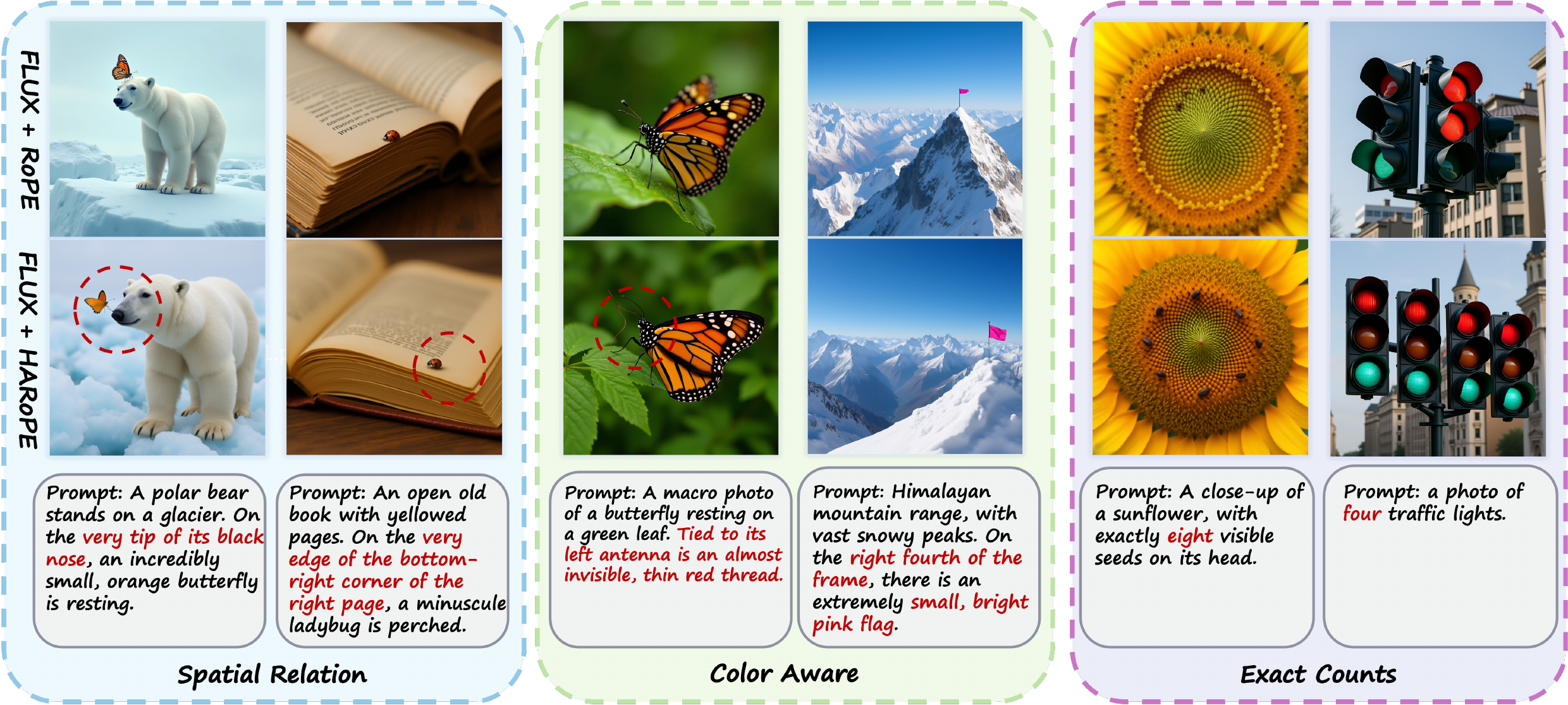}
\caption{Qualitative comparison of generated images across three fine-grained challenges: 
spatial relations (left), 
color fidelity (middle), 
and object counting (right). 
HARoPE consistently outperforms RoPE, 
adhering more faithfully to prompt specifications (instruction keywords highlighted in red).}
\label{fig:teasers}
\end{figure*}

\begin{abstract}

Transformers rely on explicit positional encoding to model structure in data. 
While Rotary Position Embedding (RoPE) excels in 1D domains, its application to image generation reveals significant limitations such as fine-grained spatial relation modeling, color cues, and object counting.
This paper identifies key limitations of standard multi-dimensional RoPE—axis-wise independence, and uniform head treatment—in capturing the complex structural biases required for fine-grained image generation. 
We propose HARoPE, a head-wise adaptive extension that inserts a learnable linear transformation parameterized via singular value decomposition (SVD) before the rotary mapping. 
% This lightweight modification enables dynamic frequency reallocation, semantic alignment of rotary planes, 
This lightweight modification enables semantic alignment of rotary planes, 
and head-specific positional receptive fields while rigorously preserving RoPE's relative-position property.
Extensive experiments on class-conditional ImageNet and text-to-image generation (Flux and SD3) demonstrate that HARoPE consistently improves performance over strong RoPE baselines and other extensions. 
The method serves as an effective drop-in replacement, offering a principled and adaptable solution for enhancing positional awareness in transformer-based image generative models.We release our code: https://github.com/Studentxll/HARoPE.
\end{abstract}
\section{Introduction}

Transformers are inherently permutation-invariant and therefore require explicit positional signals to model order and structure in sequential and spatial data~\cite{vaswani2017attention}. 
Positional embeddings meet this need by mapping position indices to vectors of the same dimensionality as token features, 
enabling the model to fuse positional and semantic information without architectural changes. 
Two broad families are widely used. 
Absolute positional encoding assigns a unique vector to each index, 
implemented either as fixed sinusoidal functions~\cite{vaswani2017attention,chen2023pixart,Peebles2022DiT} or as learnable embeddings~\cite{gehring2017convolutional}. 
Relative encodings instead inject pairwise offset information directly into the attention mechanism~\cite{shaw2018self,raffel2020exploring,dai2019transformer}, 
often improving structural bias and length generalization. 
Among these, Rotary Positional Embedding~\cite{su2024roformer,barbero2024round,kexuefm-8265,liu2023scaling} is particularly notable: 
it represents absolute positions as complex-plane rotations and induces attention scores that depend solely on relative offsets, yielding strong empirical performance and extrapolation-friendly behavior.

Despite its success in one-dimensional settings, RoPE faces fundamental challenges when extended to multi-dimensional data, especially in image generation, 
which requires fine-grained spatial relations, color-aware cues, and exact object counts (as shown in Figure~\ref{fig:teasers}). 
First, conventional designs partition feature dimensions uniformly across axes and reuse the same frequency spectrum, implicitly assuming comparable complexity, scale, and dynamics along each direction. 
This rigid allocation is often suboptimal, especially in heterogeneous domains where horizontal and vertical axes (or spatial and temporal dimensions) exhibit different frequency characteristics. 
Second, standard multi-dimensional constructions implement rotations on fixed, coordinate-indexed planes and enforce axis-wise independence through block-diagonal structures. 
These choices constrain positional encoding to predefined subspaces that may be misaligned with the model’s learned semantics and suppress cross-dimensional interactions such as diagonal, rotational, or spatiotemporal couplings. 
Third, applying a single, shared positional mapping across all attention heads overlooks their distinct roles and receptive fields, limiting the emergence of head-level specialization needed to capture multi-scale and anisotropic patterns.

Motivated by these observations, we introduce HARoPE, a head-wise adaptive rotary positional encoding mechanism that preserves RoPE’s relative-offset property while addressing the above limitations in a lightweight and modular manner. 
The key idea is to insert, immediately before the rotary mapping, a learnable linear transformation parameterized via a singular value decomposition (SVD). 
By projecting queries and keys through this SVD-based change of basis, HARoPE aligns rotary planes with semantically meaningful directions and facilitates explicit cross-axis mixing. 
Moreover, endowing each attention head with an independent SVD equips the model with specialized positional receptive fields, 
promoting complementary multi-scale behaviors.
Crucially, using the same adaptation for queries and keys preserves RoPE’s offset equivariance, encouraging that attention depends on positions only through relative differences.

Experiments on the ImageNet generation task demonstrate that HARoPE offers a simple, drop-in mechanism and obtains improved performance compared to naïve multi-dimensional RoPE and recent extensions.
When integrated into text-to-image generative models (Flux and SD3), 
HARoPE yields consistent gains, indicating that adaptive, head-wise positional rebasing complements large-scale text-to-image generative architectures.

\section{Related Works}
\paragraph{Position Embedding in Transformers.}
Transformers are permutation-equivariant and therefore require positional signals to model order and structure~\cite{vaswani2017attention}. 
Early approaches include learned absolute embeddings~\cite{chu2021conditional,gehring2017convolutional} and fixed sinusoidal encoding~\cite{vaswani2017attention,chen2023pixart,Peebles2022DiT}, the latter enabling length extrapolation. 
Relative schemes~\cite{shaw2018self,raffel2020exploring,dai2019transformer} inject pairwise distance information directly into attention, 
improving structural bias across diverse tasks.

\paragraph{RoPE and its Extensions.}
RoPE encodes absolute positions via complex-plane rotations while preserving a strict relative-offset property in attention~\cite{su2024roformer}. 
RoPE’s parameter-free, extrapolative design has driven broad adoption in large language models. 
However, its original 1D formulation is not directly aligned with the multi-dimensional inputs common in vision.
Several works extend RoPE beyond 1D: 
RoPE-ViT generalizes to images~\cite{heo2024rotary},
and MRoPE supports 2D/3D and multimodal settings~\cite{Qwen2-VL,Qwen2.5-VL}. 
Despite progress, 
common designs (i) uniformly partition feature dimensions across axes, 
(ii) enforce axis-wise independence via block-diagonal rotations, 
and (iii) apply identical positional mappings across heads—limitations that hinder alignment with learned semantics, cross-axis coupling, and head specialization. 
Complementary efforts provide broader foundations: 
RethinkRoPE~\cite{liu2025rethinkingropemathematicalblueprint} offers a systematic mathematical blueprint for higher-dimensional RoPE, 
and STRING~\cite{schenck2025learning} introduces learnable matrix generalizations. 
Building on these insights, 
we study the learnable-matrix setting and introduce a lightweight, head-specific linear adaptation via SVD parameterization that preserves RoPE's relative-offset property while enabling semantic alignment, cross-axis mixing, and per-head specialization.

\paragraph{Image Generation and Understanding.}
Diffusion-based text-to-image systems (e.g., DALL{\(\cdot\)}E~\cite{ramesh2021zero}), 
DiT~\cite{Peebles2022DiT}, 
Stable Diffusion~\cite{rombach2022high}, 
Flux~\cite{labs2025flux1kontextflowmatching}) achieve state-of-the-art generation by coupling strong text encoders with scalable Transformers. 
In visual understanding, ViT~\cite{dosovitskiy2020image,heo2021rethinking,beyer2023flexivit,li2024llama} and Swin Transformer~\cite{liu2021swin} have largely supplanted convolutional backbones by modeling long-range dependencies and enabling multimodal alignment.
In both generation and understanding tasks, effective positional encoding is critical for representing spatial and spatiotemporal structure. The proposed HARoPE method is complementary to these Transformer-based approaches. 
We demonstrate its efficacy in image generation using Flux and SD3, and in image understanding with ViT-Base.

% \begin{figure}[!t]
%     \centering
%     \includegraphics[width=1.0\textwidth]{picture/method/pipe.png}
%     \caption{
%     Comparison of standard 2D-RoPE and the proposed HARoPE. 
%     Left (a): Standard 2D-RoPE applies axis-specific block-diagonal rotations $R_m$ and $R_n$ to the query $\mathbf{q}$ and key $\mathbf{k}$, respectively, such that the resulting attention scores depend only on the relative offset n - m. 
%     Right (b): HARoPE inserts, for each attention head $h$, a learnable linear transformation $A_h$ prior to the rotary mapping. 
%     The same $A_h$ is applied to both queries and keys, ensuring that the attention scores remain a function of the relative offset $n - m$, thereby preserving RoPE's relative-position property.
%     }
%     \label{fig:compare}
% \end{figure}
\section{Methodology}
% We introduce HARoPE (Head-wise Adaptive Rotary Positional Encoding), 
% a drop-in enhancement to RoPE designed to preserve its desirable relative-position property while addressing three core limitations that arise in multi-dimensional settings:
% rigid frequency allocation, misalignment with learned semantic subspaces, 
% and uniform treatment across attention heads.
We introduce HARoPE (Head-wise Adaptive Rotary Positional Encoding), 
a drop-in enhancement to RoPE designed to preserve its desirable relative-position property while addressing two core limitations that arise in multi-dimensional settings: misalignment with learned semantic subspaces, 
and uniform treatment across attention heads.
HARoPE incorporates a lightweight, head-specific linear transformation—parameterized via a singular value decomposition—immediately before the rotary mapping.
This adaptation enables 
% (i) dynamic redistribution of positional capacity across axes, 
% (ii) semantic alignment of rotary planes and support for cross-axis interactions, 
% and (iii) specialized positional receptive fields per attention head.
(i) semantic alignment of rotary planes and support for cross-axis interactions, 
and (ii) specialized positional receptive fields per attention head.

We first review the original RoPE and a common multi-dimensional extension (Section~\ref{subsec:preliminary}), 
then detail the specific limitations of the standard approach (Section~\ref{subsec:limitation}), 
and finally present the HARoPE formulation and its properties (Section~\ref{subset:harope}).

\subsection{Preliminary: Rotary Position Embeddings}\label{subsec:preliminary}
\paragraph{One-Dimensional RoPE.}
RoPE injects position via 2D rotations applied to consecutive feature pairs. 
For a feature vector 
$\mathbf{q}\in\mathbb{R}^d$ 
at position $m$, 
define the block-diagonal rotation
% \begin{equation}
% R_m = \mathrm{diag}\!\Big(
% \begin{bmatrix}\cos(m\theta_0)&-\sin(m\theta_0)\\ \sin(m\theta_0)&\cos(m\theta_0)\end{bmatrix},
% \ldots,
% \begin{bmatrix}\cos(m\theta_{d/2-1})&-\sin(m\theta_{d/2-1})\\ \sin(m\theta_{d/2-1})&\cos(m\theta_{d/2-1})\end{bmatrix}
% \Big),
% \end{equation}
\begin{multline}
R_m = \mathrm{diag}\!\Big(
\begin{bmatrix}\cos(m\theta_0)&-\sin(m\theta_0)\\ \sin(m\theta_0)&\cos(m\theta_0)\end{bmatrix}, \\
\ldots, \\
\begin{bmatrix}\cos(m\theta_{d/2-1})&-\sin(m\theta_{d/2-1})\\ \sin(m\theta_{d/2-1})&\cos(m\theta_{d/2-1})\end{bmatrix}
\Big),
\end{multline}
with frequencies 
$\theta_i = \theta_{\mathrm{base}}^{-2i/d}$ 
(typically $\theta_{\mathrm{base}}=10000$). 
Rotated queries and keys are $\mathbf{q}'=R_m\mathbf{q}$, 
$\mathbf{k}'=R_n\mathbf{k}$. 
A key property is relative-position encoding:
\begin{equation}
(R_m\mathbf{q})^\top (R_n\mathbf{k}) \;=\; \mathbf{q}^\top R_{n-m}\mathbf{k},
\end{equation}
so attention scores depend on the offset $n-m$ only. 
Each pair $(q_{2i},q_{2i+1})$ forms a 2D plane rotated by phase $m\theta_i$, yielding a multi-frequency spectrum.

\paragraph{A Naïve Multi-dimensional Extension.}
For 2D positions $(x,y)$, a standard extension partitions the feature dimensions across axes and applies independent rotations:
\begin{equation}
R_{(x, y)} \;=\; \mathrm{diag}\big(R_x(x),\, R_y(y)\big),
\end{equation}
where $R_x(\cdot)$ and $R_y(\cdot)$ reuse the 1D spectrum. 
With $\mathbf{q}=[\mathbf{q}_x;\mathbf{q}_y]$, 
$\mathbf{k}=[\mathbf{k}_x;\mathbf{k}_y]$, 
the rotated vectors and the score can be written as
\begin{equation}
\mathbf{q}' =
\begin{bmatrix}
R_x(x) & 0\\
0 & R_y(y)
\end{bmatrix} \mathbf{q},\quad
\mathbf{k}' =
\begin{bmatrix}
R_x(x') & 0\\
0 & R_y(y')
\end{bmatrix} \mathbf{k},
\end{equation}
\begin{equation}
{\mathbf{q}'}^\top \mathbf{k}' \;=\;
\underbrace{\mathbf{q}_x^\top R_x(x)^\top R_x(x') \mathbf{k}_x}_{x\text{-axis}}
\;+\;
\underbrace{\mathbf{q}_y^\top R_y(y)^\top R_y(y') \mathbf{k}_y}_{y\text{-axis}}
\end{equation}
This separability extends to higher dimensions by adding more axis-specific blocks.

\subsection{Limitations of Naïve Multi-Dimensional RoPE}\label{subsec:limitation}
% \paragraph{Rigid Frequency Allocation.} 
% Features are split evenly across axes and each axis reuses the same spectrum 
% $\theta_i = 10000^{-2i/d}$, where the $\theta_{base}$ is manually predefined, 
% implicitly assuming equal complexity and scale across directions. 
% This assumption is often violated (e.g., temporal vs. spatial variation), leading to suboptimal capacity and frequency coverage.

\paragraph{Semantic Misalignment and Axis Independence.}
Rotations act on fixed, coordinate-indexed planes 
$(q_{0},q_{1}), (q_{2},q_{3}),\ldots$, 
irrespective of the semantic subspaces learned by the model. 
The block-diagonal structure further enforces axis-wise independence, 
suppressing explicit cross-axis interactions (e.g., diagonal or rotational couplings).

\paragraph{Head-Wise Uniformity.}
Standard RoPE injects the same positional mapping into every head, despite evidence that heads specialize in different receptive fields (local vs. long-range). 
This uniformity weakens multi-scale, head-specific positional sensitivity.

\subsection{Head-Wise Adaptive RoPE}\label{subset:harope}
% We propose HARoPE, a head-wise linear adaptation inserted immediately before the rotary mapping. 
% The adaptation learns a change of basis that 
% (i) reallocates positional capacity across axes, 
% (ii) aligns rotary planes with semantically meaningful directions and enables cross-axis coupling, 
% and (iii) allows different attention heads to specialize in distinct positional receptive fields—all while preserving RoPE’s desirable relative-position property.
We propose HARoPE, a head-wise linear adaptation inserted immediately before the rotary mapping. 
The adaptation learns a change of basis that  
(i) aligns rotary planes with semantically meaningful directions and enables cross-axis coupling, 
and (ii) allows different attention heads to specialize in distinct positional receptive fields—all while preserving RoPE’s desirable relative-position property.

\paragraph{Head-specific Linear Adaptation.}
HARoPE inserts, for each attention head $h$ with per-head dimension $d$, 
a learnable linear transform $A_h\in\mathbb{R}^{d\times d}$ immediately before the rotary map. 
We parameterize
\begin{equation}
A_h = U_h\,\Sigma_h\,V_h^\top, 
\end{equation}
where $U_h,V_h$ are orthogonal and $\Sigma_h$ is diagonal with positive entries. 
Queries and keys at positions $m$ and $n$ are mapped as
\begin{equation}
\mathbf{q}'_h = R_m\,A_h^\top\,\mathbf{q}_h,\qquad \mathbf{k}'_h = R_n\,A_h^{-1}\,\mathbf{k}_h
\end{equation}
This single linear step separates concerns: 
$V_h$ selects and mixes directions (aligning rotary planes with learned semantics), 
$\Sigma_h$ redistributes effective capacity by reweighting subspaces, 
and $U_h$ maps enriched signals back to the model’s native basis. 
Initializing $A_h=I$ recovers the baseline at step zero, and keeping singular values near 1 preserves scale.

The same $A_h$ is applied to queries and keys,
and position dependence remains confined to the rotary maps, 
HARoPE preserves strict relative-offset dependence:
% \begin{equation}
% (\mathbf{q}'_h)^\top \mathbf{k}'_h
% \;=\; (R_m A_h \mathbf{q}_h)^\top (R_n A_h \mathbf{k}_h)
% \;=\; \mathbf{q}_h^\top A_h^\top R_{n-m} A_h\,\mathbf{k}_h. 
% \end{equation}
\begin{equation}
(\mathbf{q}'_h)^\top \mathbf{k}'_h=(R_m A_h^\top \mathbf{q}_h)^\top (R_n A_h^{-1} \mathbf{k}_h)=\mathbf{q}_h^\top A_h R_{n-m} A_h^{-1}\,\mathbf{k}_h
\end{equation}
So attention scores depend on positions only through the relative offset $n-m$.

\paragraph{Multi-Dimensional Extension.}
For positions $(x_1,\ldots,x_p)$ in $p$ dimensions, 
let $R_{(x_1,\ldots,x_p)}$ be the block-diagonal rotary map formed by axis-wise rotations. 
Applying the same head-specific adaptation,
\begin{equation}
\mathbf{q}'_h \;=\; R_{(x_1,\ldots,x_p)}\, A_h^\top \mathbf{q}_h\quad
\mathbf{k}'_h \;=\; R_{(x_1,\ldots,x_p)}\, A_h^{-1} \mathbf{k}_h
\end{equation}
yields the score
\begin{equation}
(\mathbf{q}'_h)^\top \mathbf{k}'_h
\;=\; \mathbf{q}_h^\top A_h R_{(\Delta x_1,\ldots,\Delta x_p)} A_h^{-1} \,\mathbf{k}_h
\end{equation}
with $\Delta x_i = x'_i - x_i$. 
Hence, HARoPE preserves relative encoding in multi-dimensional settings while allowing learned cross-axis mixing through the dense $A_h$.

\paragraph{Initialization and stability.}
To ensure compatibility with pretrained models and stable optimization, 
we initialize $A_h=I$ via $U_h=V_h=I$ and $\Sigma_h=I$. 
Orthogonality of $U_h$ and $V_h$ can be maintained by parameterizing them through the matrix exponential of skew-symmetric matrices. 
The diagonal of $\Sigma_h$ is kept positive by softplus and regularized to remain near one to avoid exploding or vanishing norms and to preserve the variance of queries and keys.

% \textcolor{red}{To ensure compatibility with pre-trained models and stable optimization, we initialize 
% $A_h(0)=I$ 
% by setting
% $U_h=V_h=I$ 
% and 
% $\Sigma_h=I$, 
% making the adapted model identical to the baseline at step zero. 
% During training, we keep 
% $A_h$ 
% near-isometric by constraining or regularizing the singular values in $\Sigma_h$ to remain close to one, 
% which preserves the norm and variance of queries/keys and prevents gradient amplification. }
% Orthogonality of $U_h$ and $V_h$ is maintained by the SVD parameterization (e.g., via retraction or implicit re-orthogonalization), 
% ensuring angles and pairwise relationships are not distorted while allowing controlled axis reweighting through $\Sigma_h$.

\begin{figure}[!t]
\centering
\includegraphics[width=1.0\linewidth]{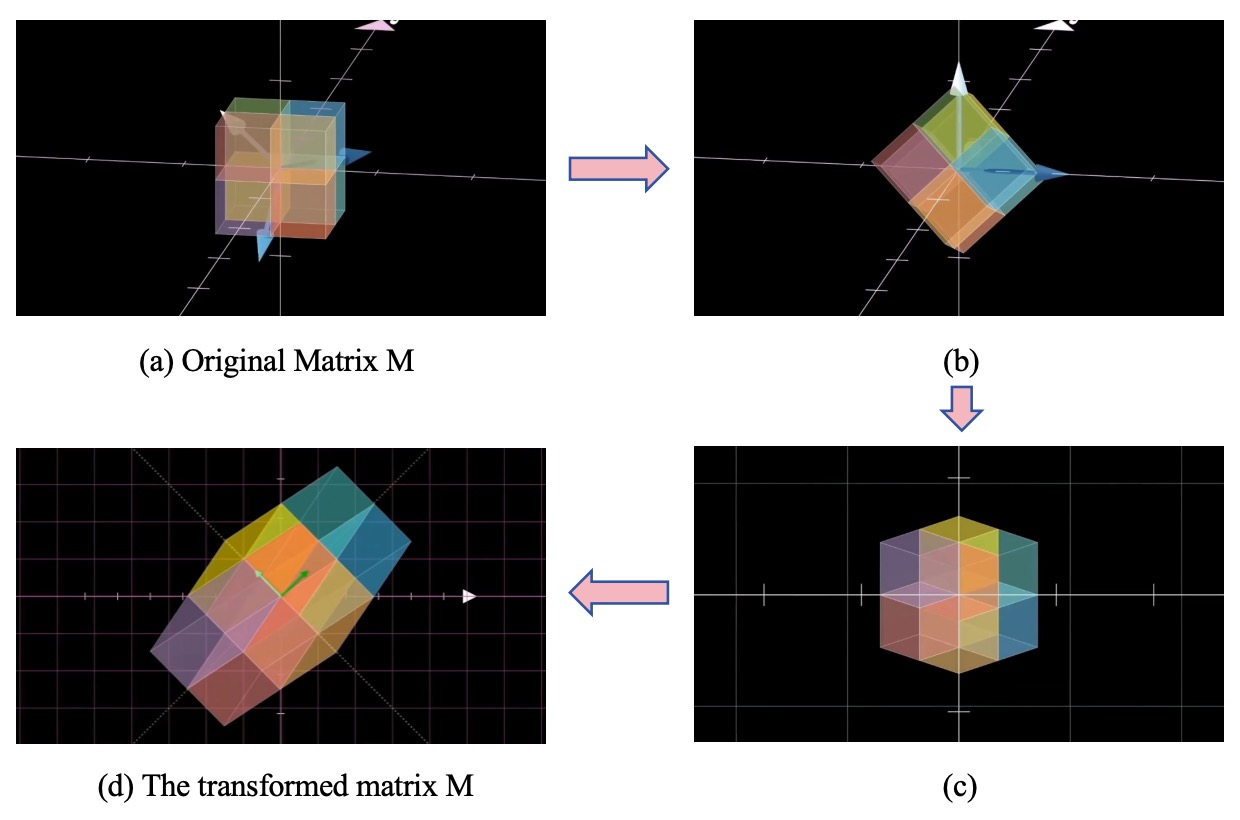}
\caption{The transformation pipeline of matrix $V_h$,$\Sigma_h$ and $U_h$.}
\label{fig:method}
\end{figure}

\paragraph{Discussion.}
HARoPE can be interpreted as learning a head-specific harmonic coordinate system: 
$V_h$ aligns rotary planes with semantically meaningful directions; 
$\Sigma_h$ modulates the effective frequency budget across these directions; 
and $U_h$ reintegrates the positionally enriched features. 
As illustrated in Figure \ref{fig:method}, we use an SVD of a $3 \times 2$ matrix as an example. Panel (a) shows the original matrix to be acted upon. After being transformed by the matrix $V_h$, its two principal components with the largest eigenvalues are aligned with the coordinate axes, with the result shown in panel (b). The $\Sigma_h$ matrix then performs axial stretching and scaling, resulting in the configuration shown in panel (c). Finally, the $U_h$ matrix rotates it onto the standard basis, yielding the final result in panel (d).
By allowing each head to specialize its positional receptive field, 
HARoPE overcomes the limitations of axis-wise independence, 
and head-wise uniformity, 
while rigorously preserving RoPE's relative-position equivariance.

\section{Experiments}
This section evaluates HARoPE across image understanding, class-conditional image generation, and text-to-image generation. We first describe the experimental protocol (architectures, datasets, baselines, and metrics), then present comparative results followed by ablations, limitations and future work discussion.

\subsection{Experimental Setups}
\paragraph{Implementation.}
We adopt standard backbones and training strategies for each task. 
For image understanding, we pretrain DeiT-Base~\cite{touvron2021training} and DeiT-Large~\cite{touvron2021training} from scratch with AdamW under the resolution of $192\times 192$, 
learning rate $5\times10^{-4}$ and a 5-epoch warmup from $1\times10^{-6}$, global batch size 2048, and 400 training epochs, then finetune 20 epochs with a fixed learning rate of $1\times 10^{-5}$ and $224\times 224$ resolution. 
For class-conditional image generation, 
we use DiT-B/2 with a constant learning rate $1\times10^{-4}$, no weight decay, batch size 256, and EMA with decay 0.9999 for evaluation, model is evaluated with checkpoint of 1M training steps.
For text-to-image generation, we fine-tune the pretrained FLUX.1-dev model for 4,000 iterations using LoRA (rank 32), 
AdamW with learning rate $2\times10^{-5}$, weight decay 0.01, and batch size 64.
All training and testing are performed on a server of eight NVIDIA H100 GPUs.

\paragraph{Dataset.}
Datasets follow conventional practice. Image understanding experiments use ImageNet at $192\times192$ and $224\times224$ with standard resize and center-crop.
For ImageNet generation, we encode images using Stable Diffusion’s VAE into $z\in\mathbb{R}^{H/8\times W/8\times 4}$ with $H\in\{128,256,512\}$. 
Text-to-image experiments with the FLUX model use the BLIP3o-60k instruction-tuning set of 60k prompt–image pairs. For SD3-medium based text-to-image generation,  we utilize the train split of the MS-COCO dataset~\cite{lin2014microsoft}.

\paragraph{Baselines.}
We compare against strong positional encoding baselines.
For image understanding, we include absolute positional embeddings (APE), 
2D-RoPE in axial and mixed forms~\citep{heo2024rotary}, 
STRING~\citep{schenck2025learning}/Rethinking RoPE~\citep{liu2025rethinkingropemathematicalblueprint}, and HARoPE. 
For class-conditional generation on ImageNet, we evaluate APE, Vanilla RoPE, 2D-RoPE (Axial), VideoRoPE~\citep{wei2025videorope}, STRING/Rethinking RoPE, and HARoPE. 
For text-to-image generation, we directly replace RoPE in FLUX and APE (Absolute Position Embedding) in SD3-medium with HARoPE  for a controlled comparison.

\paragraph{Metrics.}
For image understanding, we report Top-1 accuracy.
In class-conditional generation, we adopt ADM’s TensorFlow evaluation suite~\cite{dhariwal2021diffusion} to report FID-50K~\citep{heusel2017gans},  
Inception Score~\citep{salimans2016improved}, 
and Precision/Recall~\citep{davis2006relationship}. 
For text-to-image generation, we employ GenEval~\citep{ghosh2023geneval}, DPG-Bench~\citep{hu2024ella} and T2I-CompBench~\citep{huang2025t2i} for comprehensive assessment.

\subsection{Comparison to Existing Works}
\paragraph{Image Understanding.}
Table~\ref{tab:understanding_imagenet} summarizes DeiT-Base and DeiT-Large results. 
HARoPE achieves the best Top-1 accuracy of 83.76\%, improving upon APE by 0.25\% and surpassing the strongest RoPE variant (2D-RoPE Mixed at 83.73\%).
These gains indicate that head-wise adaptive rotary rebasing applies well to the image understanding task and is also compatible with visual understanding. 
However, in this paper, we focus on the task of fine-grained image generation, so we do not conduct extensive experimental analysis for visual understanding.

\begin{table}[!t]
\centering
%#and Vit-Large
\scriptsize  
\begin{tabular}{ccc}
\toprule
Position Embedding & Model & Top-1 Accuracy (224$\times$224)\\
% \midrule
% APE (Default) &Deit-Small & \\
% 2D-RoPE (Axial) &Deit-Small & \\
% 2D-RoPE (Mixed) &Deit-Small & \\
% STRING/Rethinking RoPE &Deit-Small & \\
% HARoPE & Deit-Small & \\
\midrule
APE (Default)           & DeiT-Base & 83.51 \\
2D-RoPE (Axial)         & DeiT-Base & 83.68 \\
2D-RoPE (Mixed)         & DeiT-Base & 83.73 \\
STRING/Rethinking RoPE  & DeiT-Base & 83.57 \\
HARoPE                  & DeiT-Base & \textbf{83.76} \\
\midrule
APE (Default)           & DeiT-Large & 84.48 \\
2D-RoPE (Axial)         & DeiT-Large & 84.60 \\
2D-RoPE (Mixed)         & DeiT-Large & 84.81\\
STRING/Rethinking RoPE  & DeiT-Large & 84.51 \\
HARoPE                  & DeiT-Large & \textbf{84.83} \\
\bottomrule
\end{tabular}
\caption{Comparison of image understanding results with different position embeddings on DeiT-Base and DeiT-Large.}
\label{tab:understanding_imagenet}
\end{table}

\paragraph{Class-Conditioned ImageNet Generation.} 
On ImageNet with DiT-B/2 (Table~\ref{tab:image_generation_dit}), 
HARoPE attains the lowest FID-50K (8.90) and the highest IS (127.01), 
while matching the strongest Precision (0.74) and achieving the best Recall (0.55). 
These results reflect improved fidelity and perceptual quality without sacrificing diversity compared to axis-separable or fixed-spectrum designs.

\begin{table}[!t]
\centering
\scriptsize  
\begin{tabular}{ccccc}
\toprule
Position Embedding&FID-50K$\downarrow$&IS$\uparrow$&Precision$\uparrow$&Recall$\uparrow$\\
\toprule
APE (Default)&11.47&110.04&0.72&0.54\\
Vanilla RoPE&9.81&121.75&0.73&0.53\\
2D-RoPE (Axial)&9.49&124.78&0.74&0.54\\
VideoRoPE&10.86&118.84&0.71&0.54\\
STRING/Rethinking RoPE&9.31&125.09&0.74&0.54\\
\midrule
HARoPE&\textbf{8.90}&\textbf{127.01}&\textbf{0.74}&\textbf{0.55}\\
\bottomrule
\end{tabular}
\caption{Comparison of image generation results on ImageNet with different position embeddings on DiT-B/2, 1M steps.}
\label{tab:image_generation_dit}
\end{table}

\paragraph{Text-to-Image Generation.}
% Replacing RoPE with HARoPE in FLUX yields consistent improvements on both GenEval and DPG-Bench (Table~\ref{tab:image_generation_flux_mmdit}). 
% On GenEval benchmark and resolution of $1024\times 1024$, the overall score increases from 0.7567 to 0.7710, while on DPG-Bench it improves from 83.26 to 83.77. 
% The relative gain is more pronounced on GenEval, which emphasizes fine-grained compositional attributes (e.g., object counting (0.9781->0.9844), colors(0.9737->1.0), spatial relations()), aligning with HARoPE’s head-wise adaptive design that enhances spatial discrimination. 
Replacing RoPE with HARoPE in FLUX.1-dev yields consistent performance gains across both GenEval and DPGBench (Table~\ref{tab:image_generation_flux_mmdit}). At the standard $1024\times1024$ resolution, HARoPE improves the GenEval Overall score from 0.757 to 0.771 and the DPGBench score from 83.26 to 83.77.

The advantages of HARoPE are more pronounced in fine-grained compositional tasks. As shown in the detailed performance on T2I-CompBench(Table~\ref{tab:dpg_results}), HARoPE consistently outperforms RoPE in spatial-related dimensions, such as 2D-Spatial (e.g., $0.307 \rightarrow 0.337$ at 1024px) and 3D-Spatial relations. This aligns with HARoPE’s head-wise adaptive design, which enhances the model's ability to discriminate complex spatial layouts by allocating specialized frequency priors to different attention heads.

Furthermore, HARoPE demonstrates superior scalability at higher resolutions. Notably, at 2048px, HARoPE significantly boosts the Position score from 0.370 to 0.455 and the Attribute score from 0.623 to 0.660. These results suggest that our head-wise rotation re-scaling effectively mitigates the frequency issues typically encountered during resolution extrapolation.

\begin{figure*}[!t]
\centering
% \fbox{\rule{0pt}{3cm}\rule{5cm}{0pt}} % 5cm宽，3cm高的空图
\includegraphics[width=1.0\textwidth]{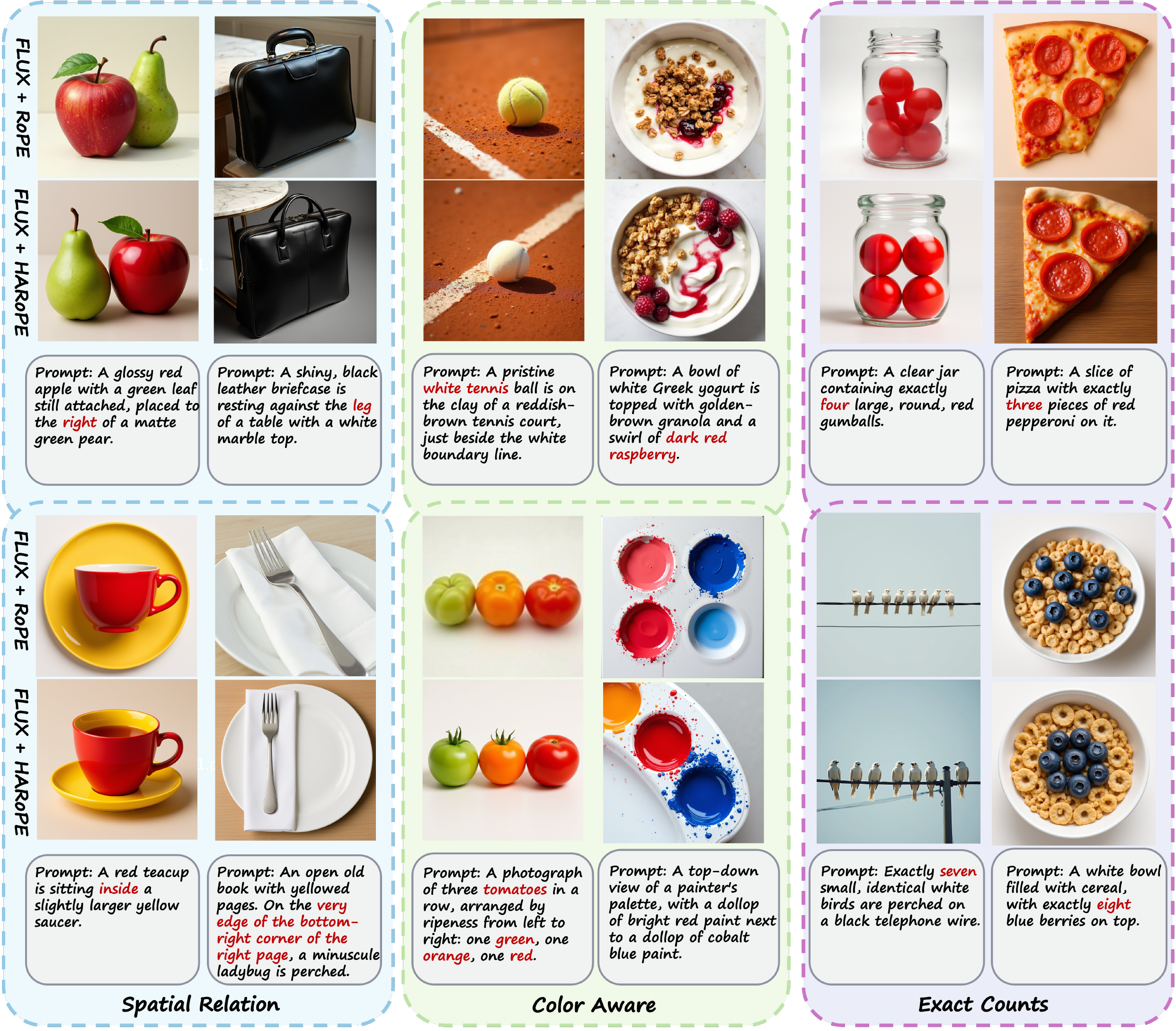}
\caption{Qualitative comparison on wild prompts, evaluating FLUX models with RoPE and HARoPE positional embeddings.}
\label{fig:flux_compare}
\end{figure*}

Applying HARoPE to SD3-medium further reduces FID from 5.35 (RoPE) to 5.22 (HARoPE), indicating improved fidelity on MS-COCO dataset. 
Qualitative comparisons are provided in Figure~\ref{fig:flux_compare}.

\begin{table}[!t]
\centering
\scriptsize
\begin{tabular}{ccc}
\toprule
Method & FID $\downarrow$ \\
\midrule
SD3-medium (APE)         & 6.34  \\
SD3-medium (RoPE)        & 5.35  \\
SD3-medium (HARoPE)      & \textbf{5.22}  \\
\bottomrule
\end{tabular}
\caption{Performance comparison of Abosultion Position Embedding and HARoPE applied to SD3-medium on 256$\times$256 resolution.}
\label{tab:image_generation_mmdit} % 只保留一个标签，并放在 caption 后面
\end{table}

% \begin{table}[!t]
% \centering
% \scriptsize
% \setlength{\tabcolsep}{2pt}
% \begin{tabular}{lcccc}
% \toprule
% Method  &\makecell{T2ICompBench \\ (spatial)} & \makecell{T2ICompBench \\ (no-spatial)} &GenEval $\uparrow$ & \makecell{DPG\\Bench $\uparrow$}  \\
% \midrule
% \multicolumn{4}{l}{\textbf{512$\times$ 512}} \\
% FLUX (RoPE)      &\textbf{0.3716}&0.1453&\textbf{0.7651}  & \textbf{82.63}   \\
% FLUX (HARoPE)    &0.3695 &\textbf{0.1477}&0.7581& 82.37   \\
% \midrule
% \multicolumn{4}{l}{\textbf{1024$\times$ 1024}} \\
% FLUX (RoPE)     &0.3704 &0.1440 &0.7567&83.26 \\
% FLUX (HARoPE)   &\textbf{0.3708}&\textbf{0.1460}&\textbf{0.7710} & \textbf{83.77}    \\
% \midrule
% \multicolumn{4}{l}{\textbf{2048 $\times$ 2048}} \\
% FLUX (RoPE)     &0.3588 & \textbf{0.1442}&0.8984 & 84.12    \\
% FLUX (HARoPE)   &\textbf{0.3633}&0.1441 &\textbf{0.9183}& \textbf{84.46} \\
% \bottomrule
% \end{tabular}
% \caption{FLUX.1-dev based text-to-image generation performance comparison of RoPE and HARoPE on three benchmarks: T2ICompBench, Geneval and DPGBench, with $512\times512$ and $1024\times1024$ resolution.}
% \label{tab:image_generation_flux_mmdit} % 只保留一个标签，并放在 caption 后面
% \end{table}

% --- 表格 1 ---
\begin{table}[!t]
\centering
\label{tab:results} % 注意：\label 放在 \caption 后面更稳妥
\resizebox{\columnwidth}{!}{ % 仅缩放表格体
\begin{tabular}{llcccccccc}
\toprule
\multirow{2.5}{*}{Res.} & \multirow{2.5}{*}{Method} & \multicolumn{7}{c}{GenEval} & \multirow{2.5}{*}{DPG.} \\ 
\cmidrule(lr){3-9}
 & & 2-Obj. & 1-Obj. & Pos. & Attr. & Count & Color & OverAll & \\ 
\midrule
\multirow{2}{*}{512} & RoPE & \textbf{0.904} & 0.988 & \textbf{0.408} & \textbf{0.606} & 0.844 & \textbf{0.843} & \textbf{0.765}& \textbf{82.63} \\
 & HARoPE & 0.894 & 0.988 & 0.388 & 0.593 & \textbf{0.866} & 0.822 & 0.758 & 82.37 \\ 
\midrule
\multirow{2}{*}{1024} & RoPE & 0.902 & 0.991 & 0.373 & 0.635 & 0.800 & 0.848 & 0.757 & 83.26 \\
 &  HARoPE & 0.902 & 0.991 & \textbf{0.395} & \textbf{0.658} & \textbf{0.828} & 0.848 & \textbf{0.771} & \textbf{83.77} \\ 
\midrule
\multirow{2}{*}{2048} & RoPE & \textbf{0.889} & 0.978 & 0.370 & 0.623 & \textbf{0.753} & 0.809 & 0.735 & 84.12 \\
 &  HARoPE & 0.881 & \textbf{0.984} & \textbf{0.455} & \textbf{0.660} & 0.734 & \textbf{0.822} & \textbf{0.756} & \textbf{84.46} \\ 
\bottomrule
\end{tabular}
} % resizebox 结束
\caption{FLUX.1-dev based text-to-image generation performance comparison of RoPE and HARoPE on two benchmarks: GenEval and DPGBench, with $512\times512$, $1024\times1024$ and $2048\times2048$ resolutions.}
\label{tab:image_generation_flux_mmdit}
\end{table}

% --- 表格 2 ---
\begin{table}[!t]
\centering
\resizebox{\columnwidth}{!}{ % 仅缩放表格体
\begin{tabular}{llccccc}
\toprule
\multirow{2.5}{*}{Res.} & \multirow{2.5}{*}{Method} & \multicolumn{5}{c}{T2I-CompBench} \\ 
\cmidrule(lr){3-7}
 & & Numeracy & 2D-Spat. & 3D-Spat. & Complex & Non-Spat. \\ 
\midrule
\multirow{2}{*}{512} & RoPE & \textbf{0.677} & \textbf{0.323} & \textbf{0.431} & \textbf{0.108} & 0.312 \\
 &  HARoPE & 0.645 & 0.297 & 0.411 & 0.101 & \textbf{0.313}\\ 
\midrule
\multirow{2}{*}{1024} & RoPE & 0.641 & 0.307 & 0.417 & 0.100 & 0.311 \\
 &  HARoPE & \textbf{0.666} & \textbf{0.337} & \textbf{0.432} & \textbf{0.101} & \textbf{0.312} \\ 
\midrule
\multirow{2}{*}{2048} & RoPE & 0.633 & 0.279 & 0.403 & 0.077 & 0.307 \\
 &  HARoPE & \textbf{0.656} & \textbf{0.322} & \textbf{0.408} & \textbf{0.094} & \textbf{0.308} \\ 
\bottomrule
\end{tabular}
} % resizebox 结束
\caption{Fine-grained evaluation on T2I-CompBench, with $512\times512$, $1024\times1024$ and $2048\times2048$ resolutions.}
\label{tab:dpg_results}
\end{table}

% \begin{figure}[!t]
% \centering
% % \fbox{\rule{0pt}{3cm}\rule{5cm}{0pt}} % 5cm宽，3cm高的空图
% \includegraphics[width=0.9\textwidth]{picture/flux_compare_2.pdf}
% \caption{Qualitative Comparison on the GenEval Benchmark, evaluating FLUX models with RoPE and HARoPE positional embeddings.}
% \label{fig:flux_compare_2}
% \end{figure}

% \begin{figure}[!t]
% \centering
% \includegraphics[width=0.9\textwidth]{picture/flux_compare_3.pdf}
% \caption{Qualitative Comparison on the GenEval Benchmark, evaluating FLUX models with RoPE and HARoPE positional embeddings.}
% \label{fig:flux_compare_3}
% \end{figure}

\subsection{Ablation Study}
\paragraph{Different Matrix Parameterizations.}
We conduct an ablation study to evaluate the impact of the matrix parameterization in HARoPE's adaptation module. 
As shown in Table~\ref{tab:multihead}, 
we compare three matrix types: normal matrices (without orthogonality constraints), orthogonal matrices, and our SVD-based parameterization. 
The baseline RoPE achieves an FID-50K of 9.49. 
Introducing a single normal matrix improves FID to 9.28, 
while orthogonal and SVD parameterizations yield 9.31 and 8.93 respectively, 
demonstrating that constrained matrix structures provide more stable optimization.

% \begin{table}[!t]
% \centering
% \scriptsize  
% \begin{tabular}{lccccc}
% \toprule
% \multicolumn{1}{l}{Position Embedding} & Training steps & \multicolumn{1}{c}{FID-50K$\downarrow$}  & \multicolumn{1}{c}{IS$\uparrow$} & \multicolumn{1}{c}{Precision$\uparrow$} & \multicolumn{1}{c}{Recall$\uparrow$} \\
% \toprule
% RoPE & DiT-B/2,1M & 9.49  & 124.78& 0.74 & 0.54 \\
% \midrule
% RoPE + normal-matrix & DiT-B/2,1M &9.28& 124.78&0.74&0.53 \\
% \midrule
% RoPE + normal-matrix + multi-head & DiT-B/2,1M & 9.03 & 126.89&0.74 &0.53 \\
% \midrule
% RoPE + orthogonal-matrix & DiT-B/2,1M &  9.31 & 125.09 & 0.74 & 0.54  \\
% \midrule
% RoPE + orthogonal-matrix + multi-head & DiT-B/2,1M &8.97  & 127.61 & 0.74 & 0.53  \\
% \midrule
% RoPE + SVD & DiT-B/2,1M & 8.93(wait to update,running) & 126.03 & 0.74& 0.54 \\
% \midrule
% % RoPE+Full-Plane orthogonal-matrix,multi-head& DiT-B,1M &H ing &  &  & &   \\
% % \midrule
% RoPE + SVD + multi-head (Ours) & DiT-B/2,1M &\textbf{ 8.90}(wait to update,running) & \textbf{127.01}& 0.74& \textbf{0.55} \\
% % \midrule
% % RoPE+norm& DiT-B,1M &  & & & &  \\
% \bottomrule
% \end{tabular}
% \caption{Quantitative comparison of different matrix settings (normal, orthogonal and SVD parameterization; with and without multi-head separate learnable matrix) on ImageNet generation task.}
% \label{tab:multihead}
% \end{table}

\begin{table}[!t]
\centering
\scriptsize
\setlength{\tabcolsep}{4pt}
\begin{tabular}{lcccc}
\toprule
\multicolumn{1}{l}{Position Embedding} & \multicolumn{1}{c}{FID-50K$\downarrow$}  & \multicolumn{1}{c}{IS$\uparrow$} & \multicolumn{1}{c}{Precision$\uparrow$} & \multicolumn{1}{c}{Recall$\uparrow$} \\
\midrule
RoPE & 9.49 & 124.78 & 0.74 & 0.54 \\
\midrule
RoPE + normal-matrix & 9.28 & 124.78 & 0.74 & 0.53 \\
\midrule
\makecell[l]{RoPE + normal-matrix\\+ multi-head} & 9.03 & 126.89&0.74 &0.53 \\
\midrule
RoPE + orthogonal-matrix & 9.31 & 125.09 & 0.74 & 0.54  \\
\midrule
\makecell[l]{RoPE + orthogonal-matrix\\+ multi-head} & 8.97  & 127.61 & 0.74 & 0.53  \\
\midrule
RoPE + SVD & 8.93 & 126.03 & 0.74& 0.54 \\
\midrule
% RoPE+Full-Plane orthogonal-matrix,multi-head& DiT-B,1M &H ing &  &  & &   \\
% \midrule
\makecell[l]{RoPE + SVD\\+ multi-head (Ours)} &\textbf{8.90} & \textbf{127.01} & 0.74 & \textbf{0.55} \\
% \midrule
% RoPE+norm& DiT-B,1M &  & & & &  \\
\bottomrule
\end{tabular}
\caption{Quantitative comparison of different matrix settings (normal, orthogonal and SVD parameterization; with and without multi-head separate learnable matrix) on ImageNet generation task.}
\label{tab:multihead}
\end{table}

\begin{figure}[!t]
\centering
\includegraphics[width=1.0\linewidth]{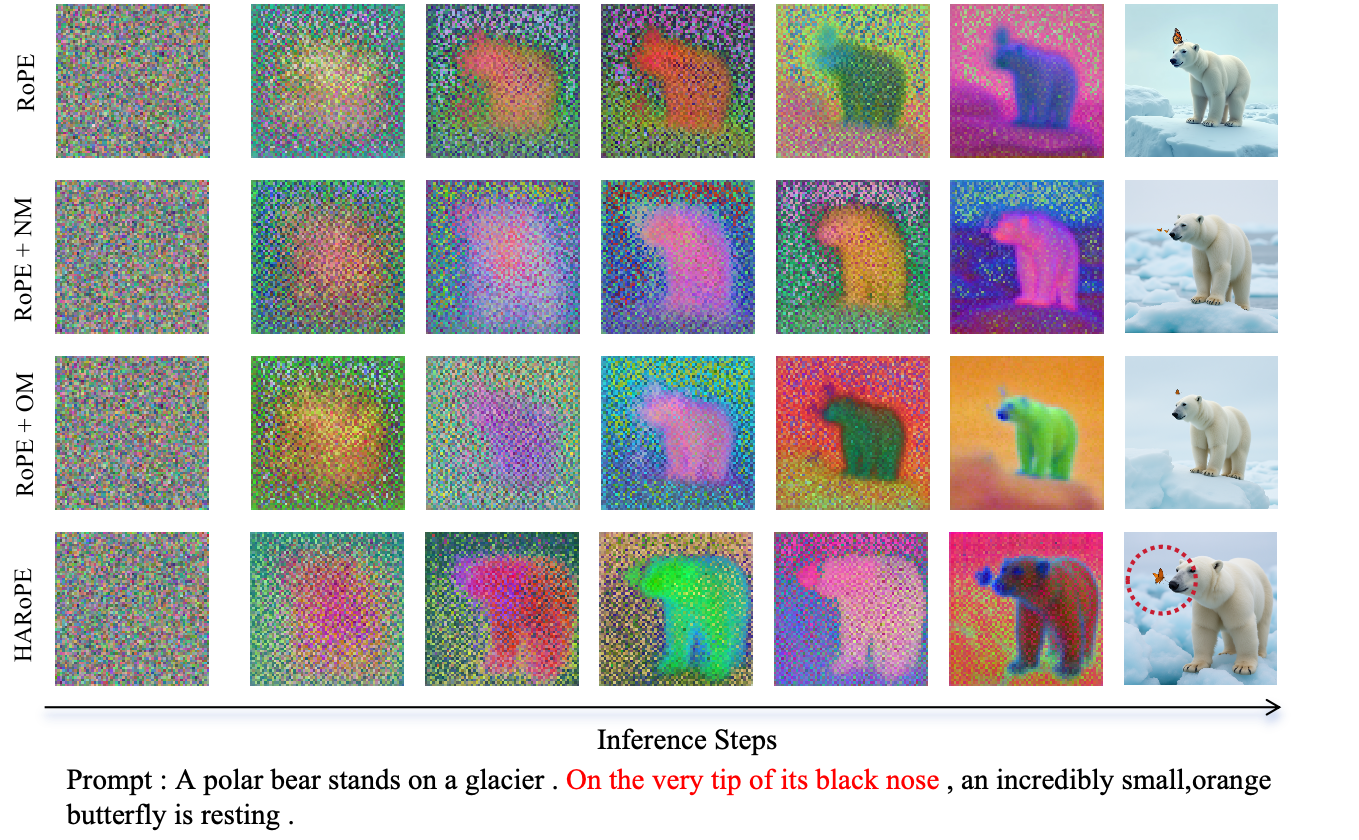}
\caption{Qualitative comparison of different matrix settings.
During the inference steps, we demonstrate the 
``NM'' denotes normal matrix, and ``OM'' denotes orthogonal matrix.}
\label{fig:multihead_pic}
\end{figure}

\begin{figure}[!t]
\centering
\includegraphics[width=1.0\linewidth]{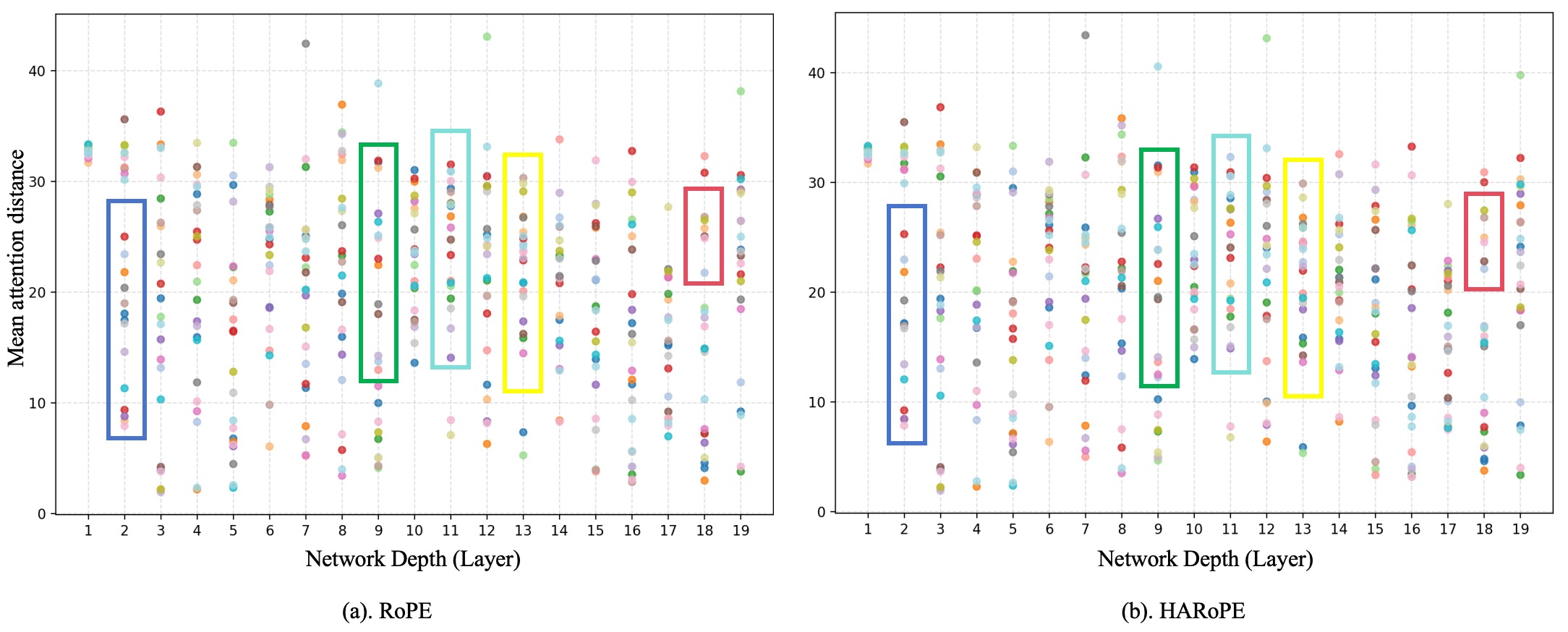}
\caption{Size of attended area by head and network depth. Attention distance was computed by averaging the distance between the query token and all other tokens, weighted
by the attention weight. Each dot shows the mean attention distance across images for one of 24
heads at one layer.}
\label{fig:multihead_compare}
\end{figure}

\paragraph{Head-wise Specialization.}
In the original RoPE, all attention heads share the same rotational frequency distribution, with different frequencies linearly assigned along the channel dimension. Under this setting, attention heads tend to learn similar spatial dependencies, resulting in a relatively concentrated distribution of attention distances within the same layer, as shown in the figure ~\ref{fig:multihead_compare}(a). In contrast, our proposed HARoPE introduces an additional transformation matrix $A_h$, which enables different attention heads to capture distinct spatial information. Consequently, the attention distance distribution across heads within the same layer becomes more uniform, as illustrated in the figure ~\ref{fig:multihead_compare}(b).

\begin{figure}[!t]
\centering
\includegraphics[width=1.0\linewidth]{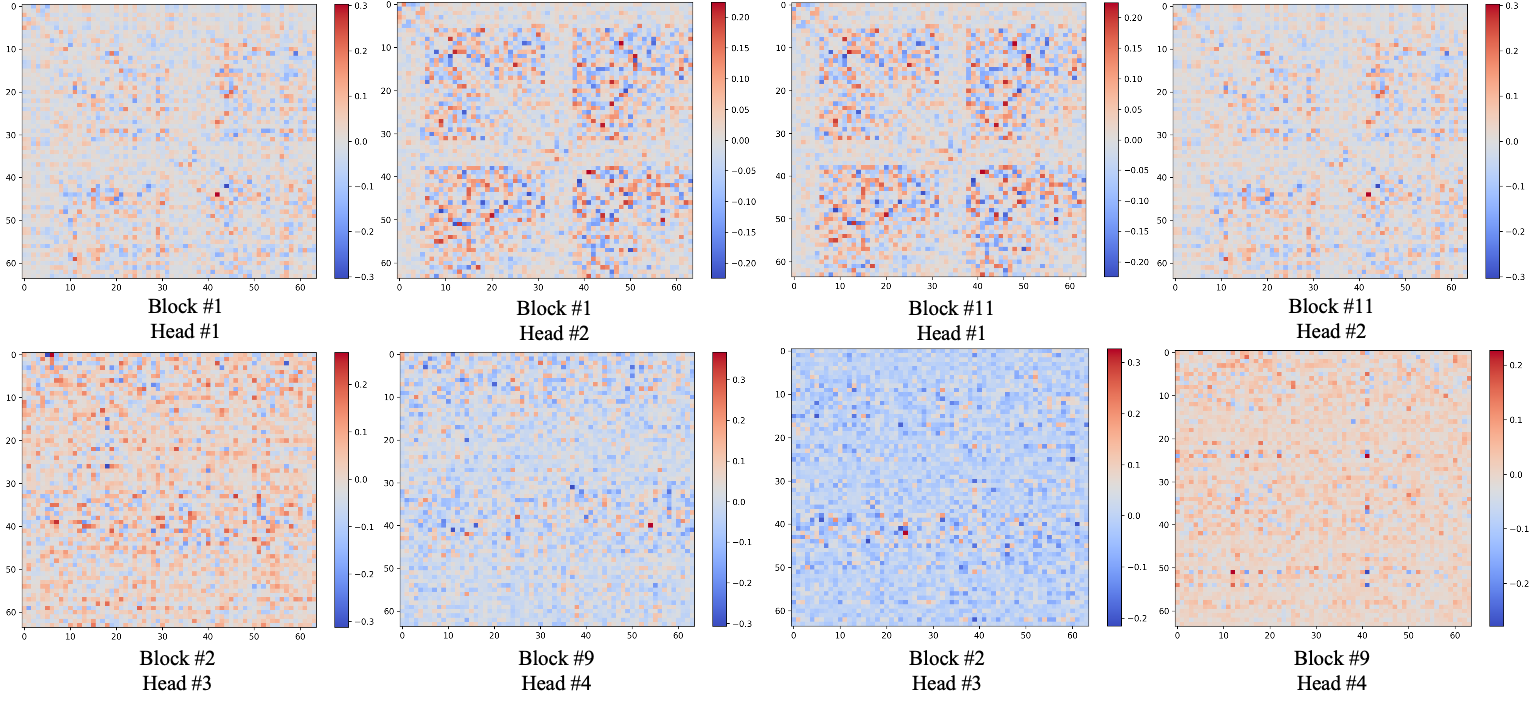}
\caption{Model weight in heatmap of different learned matrices in different attention heads and different blocks.}
\label{fig:matrix_comparison}
\end{figure}

As shown in Table~\ref{tab:multihead}, 
we extend each matrix type to be head-specific, 
and observe consistent improvements across all matrix parameterizations. 
The normal matrix with multi-head configuration reduces FID to 9.03, while the orthogonal matrix variant achieves 8.97. 
Our proposed HARoPE (RoPE + SVD + multi-head) obtains the best performance with FID-50K of 8.90 and IS of 127.01. 
This trend is corroborated in text-to-image generation with the FLUX model (Table~\ref{tab:multi_vs_single}), where the head-wise variant yields superior scores on both of GenEval and DPGBench.
To further validate this specialization, we visualize the model weight of learned transformation matrices across different attention heads and transformer blocks in Figure~\ref{fig:matrix_comparison}. 
The distinct patterns observed in the heatmaps provide empirical evidence that different heads indeed learn divergent projection strategies, aligning with the intended design of head-wise adaptive positional encoding.

\begin{table}[t!]
    \centering
    \caption{Comparing the performance of RoPE + SVD and RoPE + SVD + Multi-head on GenEval benchmark, FLUX-dev model on 1024$\times$1024 resolution.}
    \label{tab:multi_vs_single}
    \scriptsize % 保留了 \scriptsize
    
    % 注意：原来的 resizebox 是为了填满 0.53\textwidth 的 minipage。
    % 如果您希望表格保持完全相同的（比较窄的）宽度，可以取消下面这行和末尾的 } 的注释。
    % \resizebox{0.53\textwidth}{!}{%
    \begin{tabular}{c|cc}
    \toprule
    training steps & RoPE + SVD & \makecell{RoPE + SVD \\ + Multi-head (ours)} \\
    \midrule
    500  & 0.7234 & \textbf{0.7292}\\
    1000 & 0.7206 & \textbf{0.7388}\\
    2000 & 0.7311 & \textbf{0.7512}\\
    4000 & 0.7468 & \textbf{0.7710}\\
    \bottomrule
    \end{tabular}
    % } % 对应上面的 \resizebox
\end{table}

\begin{figure}[t!]
    \centering
    % 注意：宽度从 \linewidth 改为 0.45\textwidth
    % 因为它原来是 minipage (0.45\textwidth) 内部的 \linewidth
    \includegraphics[width=0.5\textwidth]{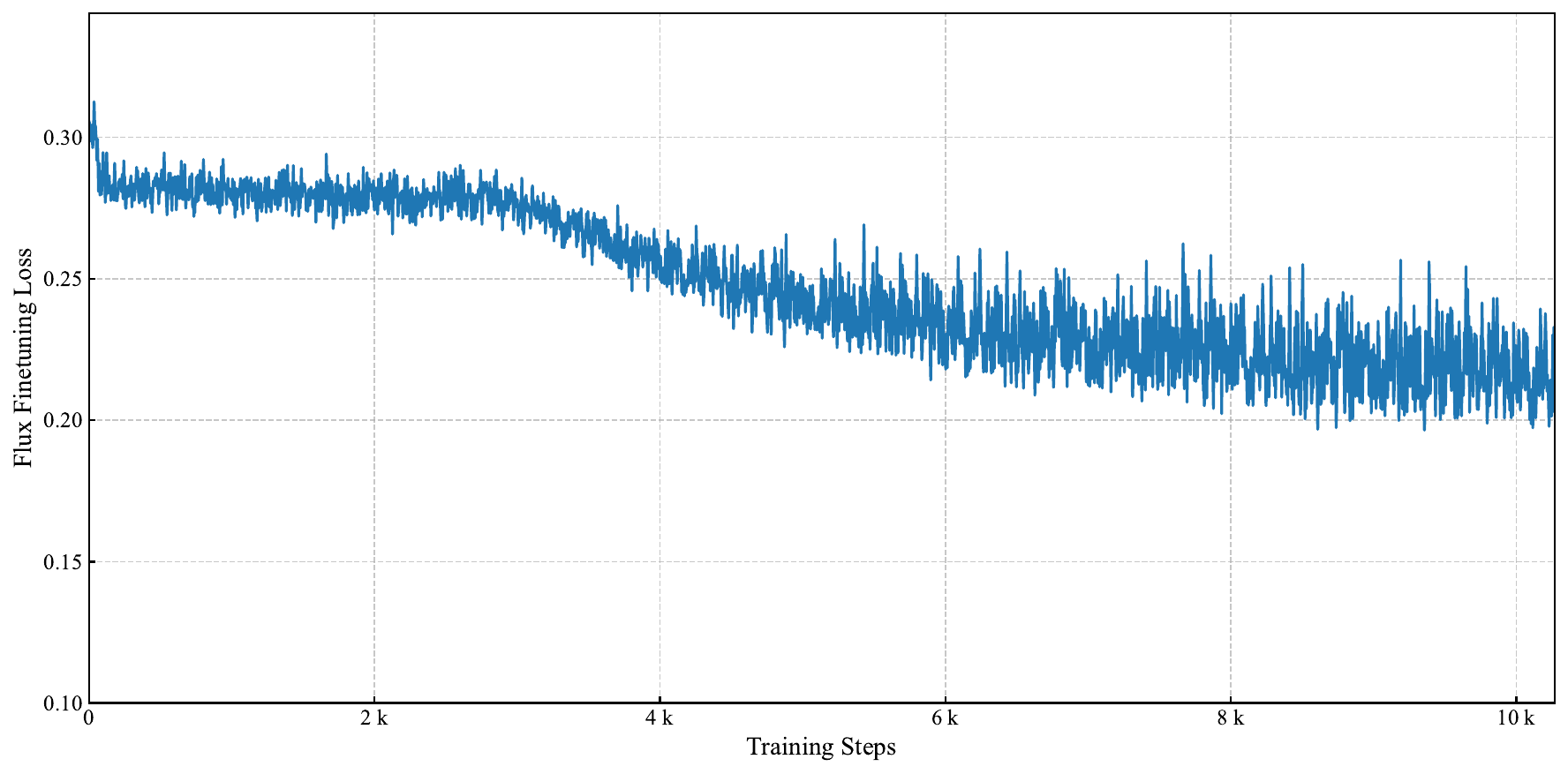}
    \caption{
    % Illustration of FLUX model training loss during further finetuning. This loss curve shows that the whole training process is stable and converges progressively after adopting our proposed HARoPE.
    Training loss of the FLUX model during finetuning, showing stable and progressive convergence with HARoPE.
    }
    \label{fig:flux_ft_loss}
\end{figure}

% \begin{figure*}[t!]
%     \centering
%     % 左边表格
%     \begin{minipage}{0.53\textwidth}
%         \centering
%         \scriptsize
%         \resizebox{\linewidth}{!}{%
%         \begin{tabular}{c|cc}
%         \toprule
%         training steps & RoPE + SVD & \makecell{RoPE + SVD \\ + Multi-head (ours)} \\
%         \midrule
%         500  & 0.7234 & \textbf{0.7292}\\
%         1000 & 0.7206 & \textbf{0.7388}\\
%         2000 & 0.7311 & \textbf{0.7512}\\
%         4000 & 0.7468 & \textbf{0.7710}\\
%         \bottomrule
%         \end{tabular}
%         }
%         \caption{Comparing the performance of RoPE + SVD and RoPE + SVD + Multi-head on GenEval benchmark, FLUX model, 1024$\times$1024 resolution.}
%         \label{tab:multi_vs_single}
%     \end{minipage}
%     \hfill
%     % 右边图片
%     \begin{minipage}{0.45\textwidth}
%         \centering
%         \includegraphics[width=\linewidth]{picture/training_loss.pdf}
%         \caption{
%         % Illustration of FLUX model training loss during further finetuning. This loss curve shows that the whole training process is stable and converges progressively after adopting our proposed HARoPE.
%         Training loss of the FLUX model during finetuning, showing stable and progressive convergence with HARoPE.
%         }
%         \label{fig:flux_ft_loss}
%     \end{minipage}
% \end{figure*}

\paragraph{Different Image Resolutions.}
We evaluate the robustness of HARoPE across multiple image resolutions to assess its scalability. 
As summarized in Table~\ref{tab:diff_res_metrics}, HARoPE is applied to DiT-B/2 models trained for class-conditional generation at resolutions of $128\times128$, $256\times256$, and $512\times512$. 
The results demonstrate that HARoPE consistently outperforms both Absolute Positional Embeddings and the standard RoPE baseline across all resolutions, achieving the best FID and IS scores.
Furthermore, as shown in Table~\ref{tab:image_generation_flux_mmdit}, when integrated into the large-scale FLUX model for text-to-image generation at a high resolution of $1024\times1024$, HARoPE again consistently yields improved performance on both the GenEval and DPG-Bench metrics compared to the original RoPE.

\begin{table}[!t]
\centering
\scriptsize
\setlength{\tabcolsep}{2pt}
\begin{tabular}{cccccc}
\toprule
{Model} & Position Embedding & \multicolumn{1}{c}{FID-50K$\downarrow$}  & \multicolumn{1}{c}{IS$\uparrow$} & \multicolumn{1}{c}{Precision$\uparrow$} & \multicolumn{1}{c}{Recall$\uparrow$} \\
\midrule
DiT-B/2, 128$\times$128& Absolution Embedding & 16.43   &58.30 &0.61&0.56\\
DiT-B/2, 128$\times$128& RoPE &  14.32 & 66.13 & 0.62 &0.57 \\
DiT-B/2, 128$\times$128& HARoPE & \textbf{13.73}   & \textbf{68.14}&\textbf{0.63}&\textbf{0.57} \\
\midrule
DiT-B/2, 256$\times$256& Absolution Embedding &11.47 & 110.04&0.72 & 0.54\\
DiT-B/2, 256$\times$256& RoPE &  9.49  & 124.78& 0.74 & 0.54\\
DiT-B/2, 256$\times$256& HARoPE &\textbf{ 8.90} & \textbf{127.01}& 0.74& \textbf{0.55} \\
\midrule 
DiT-B/2, 512$\times$512& Absolution Embedding &18.28&81.62 &0.77&0.53\\
DiT-B/2, 512$\times$512& RoPE & 14.57  &95.41 &0.79&0.53\\
DiT-B/2, 512$\times$512& HARoPE &  \textbf{14.36}  & \textbf{96.25} & \textbf{0.80} & \textbf{0.54} \\
\bottomrule
\end{tabular}
\caption{Image generation results on different image resolutions with our proposed HARoPE.}
\label{tab:diff_res_metrics}
\end{table}

\definecolor{DarkGreen}{rgb}{0.0, 0.5, 0.0}

\begin{table}[!t]
\centering
\resizebox{\columnwidth}{!}{ % 强制适应单栏宽度
\begin{tabular}{@{}llcccc@{}}
\toprule
\textbf{Model} & \textbf{Method} & \textbf{Params (M)} & \textbf{FLOPs (G)} & \textbf{FID} $\downarrow$ & \textbf{GenEval} $\uparrow$ \\ \midrule
\multirow{3}{*}{\shortstack[l]{DiT-B/2 \\ \scriptsize (256)}} & RoPE & 2.36 & 0.71 & 9.49 & -- \\
 & RoPE+SVD & 2.46 {\color{DarkGreen}\tiny (+4.1\%)} & 0.74 {\color{DarkGreen}\tiny (+4.1\%)} & 9.46 & -- \\
 &HARoPE & 2.46 {\color{DarkGreen}\tiny (+4.1\%)} & 0.74 {\color{DarkGreen}\tiny (+4.1\%)} & \textbf{8.90} & -- \\ \midrule
\multirow{3}{*}{\shortstack[l]{FLUX \\ \scriptsize (512)}} & RoPE & 28.32 & 43.49 & -- & \textbf{0.765} \\
 & RoPE+SVD & 29.11 {\color{DarkGreen}\tiny (+2.7\%)} & 44.80 {\color{DarkGreen}\tiny (+3.0\%)} & -- & 0.745 \\
 & HARoPE & 29.11 {\color{DarkGreen}\tiny (+2.7\%)} & 44.80 {\color{DarkGreen}\tiny (+3.0\%)} & -- & 0.758 \\ \midrule
\multirow{3}{*}{\shortstack[l]{FLUX \\ \scriptsize (1024)}} & RoPE & 28.32 & 130.5 & -- & 0.757 \\
 & RoPE+SVD & 29.11 {\color{DarkGreen}\tiny (+2.7\%)} & 134.1 {\color{DarkGreen}\tiny (+2.8\%)} & -- & 0.756 \\
 &  HARoPE & 29.11 {\color{DarkGreen}\tiny (+2.7\%)} & 134.1 {\color{DarkGreen}\tiny (+2.8\%)} & -- & \textbf{0.771} \\ \midrule
\multirow{3}{*}{\shortstack[l]{FLUX \\ \scriptsize (2048)}} & RoPE & 28.32 & 478.4 & -- & 0.735 \\
 & RoPE+SVD & 29.11 {\color{DarkGreen}\tiny (+2.7\%)} & 491.7 {\color{DarkGreen}\tiny (+2.7\%)} & -- & 0.731 \\
 &  HARoPE & 29.11 {\color{DarkGreen}\tiny (+2.7\%)} & 491.7 {\color{DarkGreen}\tiny (+2.7\%)} & -- & \textbf{0.756} \\ \bottomrule
\end{tabular}
}
\caption{Analysis of efficiency and performance. HARoPE introduces negligible overhead compared to the parameter-matched RoPE+SVD baseline while significantly improving performance, particularly at high resolutions.}
\label{tab:efficiency_analysis}
\end{table}

\begin{table}[!t]
    \centering
    \begin{tabular}{lccccc}
        \toprule
        $\theta_{base}$ & 5,000 & 10,000 & 30,000 & 50,000 & HARoPE \\
        \midrule
        FID$\downarrow$ & 20.45 & 20.77 & 20.83 & 20.76 & \textbf{19.01} \\
        \bottomrule
    \end{tabular}
\caption{Frequency tuning on DiT-B/2 (100k steps).}
\label{tab:fid_comparison}
\end{table}
\paragraph{Efficiency and Training Stability.}
% As shown in Table~\ref{tab:image_generation_flux_mmdit}, the TFLOPS introduced by the learnable matrices of HARoPE during inference can be very small compared to the entire model.
% The training process remains stable, as evidenced by the smooth and convergent loss curves during the fine-tuning of large models like FLUX (Figure~\ref{fig:flux_ft_loss}).
To isolate structural benefits from increased capacity, we evaluate “RoPE+SVD”, a parameter matched baseline inserting identical SVD layers post projection. As demonstrated in Table \ref{tab:efficiency_analysis}, despite having the same parameter count and computational cost as HARoPE, RoPE+SVD yields negligible improvements over the vanilla RoPE, and even suffers from performance degradation in certain GenEval metrics (e.g., $0.735 \rightarrow 0.731$ at 2048px). More importantly, the performance gap between the two becomes significantly more pronounced at higher resolutions ($1024\times1024$ and $2048\times2048$), where HARoPE consistently maintains its leading edge. These findings empirically confirm that the efficacy of our method stems from the specialized semantic alignment of rotary planes, which cannot be replicated by simply increasing the number of trainable parameters.

To demonstrate that the performance gains achieved by HARoPE cannot be obtained through simpler head-dependent frequency tuning, we conduct the experiments shown in Table \ref{tab:fid_comparison}. While head-dependent frequency tuning merely rescales the basis angles of rotary axes for each attention head—essentially performing a global rescaling of coordinate axes that fails to model complex dependencies across different feature dimensions,HARoPE introduces a learnable transformation that identifies and optimizes the optimal rotation planes through feature recombination, thereby effectively capturing cross-dimensional interactions. As shown in Table \ref{tab:fid_comparison}, HARoPE substantially outperforms the best frequency-tuned baseline (19.01 vs. 20.45 FID), validating the effectiveness of its design.

The training process remains stable, as evidenced by the smooth and convergent loss curves during the fine-tuning of large models like FLUX (Figure~\ref{fig:flux_ft_loss}).

\section{Conclusion}
Standard multi-dimensional extensions of RoPE face limitations in handling complex data like images, due to their rigid axis-wise feature partitioning, 
fixed rotation planes misaligned with semantic subspaces, 
and uniform application across attention heads. 
To overcome these issues, we introduced HARoPE, a head-wise adaptive rotary positional encoding that enhances RoPE through a lightweight, 
learnable linear transformation applied before the rotary mapping. 
Parameterized via singular value decomposition, 
this adaptation enables dynamic redistribution of positional capacity, 
semantic alignment of rotary planes with support for cross-axis interactions, 
and specialized positional receptive fields per attention head—all while preserving RoPE’s strict relative-position encoding property. 
Extensive experiments on image understanding, 
class-conditional generation, 
and text-to-image synthesis demonstrate that HARoPE consistently outperforms existing positional encoding methods, 
confirming its effectiveness as a drop-in improvement for transformer-based generative models. 
These results highlight the value of adaptive, head-wise positional reasoning in capturing fine-grained structural and semantic patterns image generative models.

\section{Limitations and Future Works}
While HARoPE demonstrates consistent improvements across image understanding and generation tasks, 
this work has certain limitations that merit discussion. 
Our evaluation is primarily confined to the image domain due to our computational constraints; 
the generalizability of the approach to other multi-dimensional data modalities, such as video, audio, or 3D content, remains an open question for empirical validation.

Another consideration is the static nature of the learned transformation matrices, which are fixed after training. 
Although the head-wise specialization is beneficial, the adaptation process is not input-conditional. 
Exploring dynamic transformations that can adapt based on input content or evolve during inference could further enhance the flexibility and performance of the positional encoding mechanism.

\section*{Acknowledgments}
This work was supported in part by the Shanghai Municipal Commission of Economy and Informatization (No.~2025-GZL-RGZN-BTBX-01011).
{
    \small
    \bibliographystyle{ieeenat_fullname}
    \bibliography{iclr2026_conference}
}

% WARNING: do not forget to delete the supplementary pages from your submission 
% \input{section/X_suppl}

\end{document}